\pdfoutput=1

\documentclass[11pt]{article}

\usepackage[]{acl}

\usepackage{times}
\usepackage{latexsym}

\usepackage{inconsolata}

\usepackage{algorithm2e}

\usepackage{graphicx}
\usepackage{subfig}
\usepackage{multirow}

\usepackage[T1]{fontenc}

\usepackage[utf8]{inputenc}

\usepackage{microtype}

\usepackage{inconsolata}

\title{Analyzing Sentiment Polarity Reduction in News Presentation through Contextual Perturbation and Large Language Models}

\author{Alapan Kuila, Somnath Jena, Sudeshna Sarkar, Partha Pratim Chakrabarti \\
        IIT Kharagpur, India \\ \texttt\{{alapan.cse, somnathjena.2011}\}@gmail.com; \texttt\{{sudeshna, ppchak}\}@cse.iitkgp.ac.in} 

\begin{document}
\maketitle

\newcommand{\mycomment}[1]{}
\begin{abstract}
In today's media landscape, where news outlets play a pivotal role in shaping public opinion, it is imperative to address the issue of sentiment manipulation within news text. News writers often inject their own biases and emotional language, which can distort the objectivity of reporting. This paper introduces a novel approach to tackle this problem by reducing the polarity of latent sentiments in news content. Drawing inspiration from adversarial attack-based sentence perturbation techniques and a prompt-based method using ChatGPT, we employ transformation constraints to modify sentences while preserving their core semantics. Using three perturbation methods—replacement, insertion, and deletion—coupled with a context-aware masked language model, we aim to maximize the desired sentiment score for targeted news aspects through a beam search algorithm. Our experiments and human evaluations demonstrate the effectiveness of these two models in achieving reduced sentiment polarity with minimal modifications while maintaining textual similarity, fluency, and grammatical correctness. Comparative analysis confirms the competitive performance of the adversarial attack-based perturbation methods and prompt-based methods, offering a promising solution to foster more objective news reporting and combat emotional language bias in the media.
\end{abstract}

\section{Introduction}

News media plays a crucial role in building public opinion on different socio-political issues and events by providing information regarding relevant facts and events. In an ideal case, the news outlets should provide the readers with correct and objective information without any bias or slant. However, news writers intentionally or unintentionally insert their own prejudice or perspectives that reflect how they view reality and what they assume to be the truth in the news broadcasts. While writing news reports, they often use emotionally charged words like subjective adjectives and sensational verbs in news text to manipulate readers' perceptions. These emotion triggering words induce sentiment bias in news reporting, which affects objective journalism. \mycomment{Therefore, reducing this latent sentiment is essential to provide balanced news reporting and more objective news stories. In order to make the news articles more objective, it is essential to reduce the intensity of this latent sentiment for generating balanced news reporting.}It is crucial to lessen the intensity of the hidden sentiment in order to provide balanced news reporting and more objective news stories. Hence, we need to rephrase the sentences while preserving the semantics to reduce the polarity of the latent sentiments.

\begin{table}[]
\fontsize{6.5}{10}\selectfont
\begin{tabular}{|c|c|}
\hline
\textbf{Input Sentence}                                                                                 & \textbf{Rephrased Sentence}                                                                                     \\ \hline
\begin{tabular}[c]{@{}c@{}}The action of the current \\ government hurt the soul of India.\end{tabular} & \begin{tabular}[c]{@{}c@{}}The action of the current government\\ impacted the spirit of India.\end{tabular} \\ \hline \hline
\begin{tabular}[c]{@{}c@{}}He alleged that the bill \\ ignored Indian Muslims.\end{tabular}             & \begin{tabular}[c]{@{}c@{}}He claimed that the bill \\ overlooked Indian Muslims.\end{tabular}                  \\ \hline \hline
\begin{tabular}[c]{@{}c@{}}Demonetisation and GST\\ will boost the economy.\end{tabular}                & \begin{tabular}[c]{@{}c@{}}Demonetisation and GST\\ may improve the economy.\end{tabular}                       \\ \hline

\end{tabular}
\caption{Instances of rewriting the sentences to lessen the polarity of implicit sentiment preserving the meaning.}
\label{example1}
\end{table}

Consider the instances of news sentences shown in the Table~\ref{example1}. The first sentence expresses a strong negative sentiment towards \textit{government action}. By replacing the terms like `hurt' and `soul' with `impacted' and `spirit' respectively, we can convey the same information with less intensity compared to the original sentence. The second sentence has a negative assessment regarding \textit{the bill}. In this case, the words like `alleged' and `ignored' induce extreme negativity. We toned down the polarity of the sentence by replacing those words with `claimed' and `overlooked'. The third sentence describes \textit{Demonetization} and \textit{GST}, two Indian government policies with positive sentiment. We can lessen the strength of the positive polarity by using the terms `may' and `improve' instead of `will' and `boost'. Overall, the problem revolves around rewriting the sentences such that the intensity of the sentiment toward a specific aspect is reduced and is represented neutrally as much as possible, preserving the semantic meaning.

 There exist some works which address a similar type of problem of rewriting the sentence while focusing on neutralizing gender bias, age bias, and demographic bias. ~\cite{He2021DetectAP} propose a two-step framework for neutralizing sentences via rewriting. In the first step, they identify the parts of the input sentence that reveals the target attribute(age/ gender/ origin bias) and mask those words. In the second step, they regenerate the complete sentence by unmasking the sentence such that the output sentence does not reveal the target attribute. During regeneration, they use a gradient-based inference method to make the output sentence attribute-neutral. Their work is inspired by PPLM~\cite{Dathathri2019PlugAP} that designs a gradient-based inference mechanism for controlled text generation from transformer-based language model. Though this method can regenerate fluent sentences that are neutral to the target attributes, it requires a fine-tuning step that may be computationally intensive and resource-demanding. 
 
 ~\cite{Pryzant2019AutomaticallyNS} try to neutralize the subjective bias in the sentence by suggesting edits that would make the sentence more neutral. They also provide a parallel corpus of biased and corresponding unbiased sentence pairs from Wikipedia edits. For the neutralization task, they propose a pair of sequence-to-sequence learning algorithms. However, the scope of this work is restricted to single-word modifications. 
 
 Some authors ~\cite{Lample2018MultipleAttributeTR, He2020APF} employ unsupervised style transfer methods to address the task of sentiment transfer where sentences are paraphrased such that it induces a different sentiment while preserving the original content. These works employ several techniques like VAEs~\cite{prabhumoye-etal-2018-style, shen2017}, adversarial network learning~\cite{Fu2017StyleTI}, sequence to sequence learning~\cite{Krishna2020ReformulatingUS} for unsupervised style transfer for manipulating sentiment. However, all these approaches require explicit sentiment labels or parallel corpus containing positive-negative or biased-unbiased sentences. In real-life scenarios, accumulating domain-specific parallel corpus or explicit sentiment-labeled sentences is a cumbersome task. Besides, previous works modified the sentences to change the overall sentiment of the text e.g., to make the sentence more positive, more neutral, or more politically slanted. But the sentences in news articles may contain more than one news aspect. For example, consider the following sentence 
 \begin{center}
 ``The government's successful rollout of testing and vaccination campaigns has been a significant step forward in combating the pandemic, but the persistent issues of healthcare inequality and insufficient support for healthcare professionals require sustained attention and action.''
 \end{center}

This sentence presents \textit{government action} positively, whereas the overall \textit{healthcare situation} is depicted negatively. Take a look at another sentence,
\begin{center}
``However, the government's handling of the protests, including the imposition of internet shutdowns and arrests of protest leaders, has been viewed as suppressive and lacking in efforts to engage in constructive dialogue with the farmers.''
\end{center}
In this sentence, \textit{government action} is depicted negatively but mentions \textit{farmers' protest} neutrally. These sentences contain more than one aspect with different sentiments. Nevertheless, the existing works do not address the problem of transferring the sentiment of one aspect, keeping the sentiment of another aspect unchanged~\cite{jin-etal-2022-deep}.  

In this paper, our primary objective is to address the novel problem of sentence rewriting, focusing on the modification of sentiment intensity pertaining to the news aspect within the sentence, all while preserving its implicit meaning. To achieve this, we employ two distinct approaches for sentence modification: the first is an adversarial attack-based perturbation method, and the second leverages prompt engineering with a large language model, ChatGPT. Our versatile approach adheres to a set of transformation constraints, facilitating the effective modification of the input news text. The proposed adversarial attack based transformation method involves perturbing the input sentence by masking a portion of the text and replacing the mask with an alternative using a context-aware, pre-trained masked language model. Within our experimentation, we explore three types of perturbations: 1) \textbf{replace}, involving the substitution of a token with a new one; 2) \textbf{insert}, entailing the addition of a new token; and 3) \textbf{delete}, which involves the removal of a token. To support these transformations, we employ a beam search method that navigates through a set of candidate perturbations to attain the desired objective. In our second approach, we harness the capabilities of a large language model, ChatGPT, by employing effective prompt engineering. This strategic combination allows us to proficiently rewrite sentences, achieving the desired sentiment modification while carefully preserving the essential semantic content. This approach not only benefits from the power of a large language model but also leverages the precision of prompt engineering, enhancing our ability to navigate the complexities of sentiment reduction in news content.

Our main contributions are threefold:

\begin{itemize}
    \item Our work address the novel problem of sentence rewriting, focusing on the modification of sentiment intensity pertaining to the news aspect within the sentence, all while preserving its implicit meaning.

    \item To achieve this, we employ two distinct approaches for sentence modification: the first is an adversarial attack-based perturbation method, and the second leverages prompt engineering with a large language model, ChatGPT. These two methodologies offer a robust and flexible means to achieve our goal of neutralizing sentiment in news text.

    \item Our extensive sentence rewriting efforts have resulted in the creation of a valuable parallel corpus~\footnote{\url{https://github.com/alapanju/NewsPolarityReduction }}, encompassing original sentences and their corresponding neutral sentiment counterparts, which represents a significant contribution to the research community.

\end{itemize}

\mycomment{
Our contribution can be summarized as follows:
\begin{itemize}
    \item We have developed a pretrained masked language model-based framework for the style transfer of news text that will change the sentiment of news representation while preserving its semantics. Our proposed framework uses a beam search method to search through candidate transformations of the input text.
    
    \item We have performed various ablation studies by changing the components and hyper-parameters of our framework to study their effectiveness in the task of style transfer on our annotated corpus.
\end{itemize}
}

\section{Methodology}
\subsection{Overview and Task Definition}
\label{task_def}

We propose a method to mitigate the sentiment polarity in the news by rewriting the sentences. Our principle goal is to neutralize the news aspect-specific sentiment polarity as a way to support objective journalism. Therefore, we focus on maximizing the neutral probability of the targeted news aspects in the sentences. We also demonstrate the efficacy of our method to perturb the sentence for maximizing the positive and negative sentiment as well, which reflects its robustness. Table~\ref{pos_neg_ex} demonstrates an example of sample modifications applied on an input sentence for maximizing three target sentiments. \mycomment{We observe that the obtained outputs are visually plausible, though in the case of negative style transfer, the information is lost/changed in describing the move as "drastic". This is natural, since it is quite difficult for even a human to modify the input sentence which is originally positive for the aspect \textit{note ban}(demonetization) to be negative.}

\begin{table*}[!ht]
\fontsize{6}{7}\selectfont
\resizebox{\textwidth}{!}{%
\begin{tabular}{l|l|l|l|l}
\hline
\textbf{Input Sentence} & \textbf{\begin{tabular}[c]{@{}l@{}}Original \\ Sentiment\end{tabular}} & \textbf{\begin{tabular}[c]{@{}l@{}}Target \\ Sentiment\end{tabular}} & \textbf{Modified sentence} & \textbf{\begin{tabular}[c]{@{}l@{}}Change in \\ sentiment \\ prediction\end{tabular}} \\ \hline
\multirow{3}{*}{\begin{tabular}[c]{@{}l@{}}The decision to declare Rs 500 and \\ Rs 1000 notes invalid from this midnight \\ is a welcome move. (Topic: Demonetization, Aspect: note ban)\end{tabular}} & \multirow{3}{*}{positive (0.98)} & positive & \begin{tabular}[c]{@{}l@{}}The \colorbox{yellow}{announcement} to make Rs 500 and \\ Rs 1000 notes invalid from this midnight \\ is a \colorbox{yellow}{brave} move.\end{tabular} & 0.01 \\ \cline{3-5} 
 &  & neutral & \begin{tabular}[c]{@{}l@{}}The decision to declare Rs 500 and \\ Rs 1000 \colorbox{yellow}{denominations} invalid from this \\ midnight is a \colorbox{yellow}{significant} move.\end{tabular} & 0.59 \\ \cline{3-5} 
 &  & negative & \begin{tabular}[c]{@{}l@{}}The \colorbox{yellow}{action} to declare Rs 500 and \\ Rs 1000 notes invalid from this midnight \\ is a \colorbox{yellow}{drastic} move.\end{tabular} & 0.91 \\ \hline
\end{tabular}%
 }
\caption{Example output of sentence perturbation method for various target sentiments. Yellow words show the modifications done. Here the aspect was \textit{note ban} (Topic: demonetization)}.
\label{pos_neg_ex}
\end{table*}

\mycomment{Our main objective is to neutralize the news-aspect specific sentiment polarity reduction as a way to support objective journalism. Therefore, this paper mainly focus on maximizing the neutral probability of the targeted news-aspect in the sentence i.e., maximize $P(S',a_t)$.} 
We will now present a formal problem definition. A news article contains a sentence $S = w_1w_2...w_n$ of $n$ words that depicts a news aspect $a_t$. The task is to produce a modified sentence $S'$ such that $P_s(S',a_t)$ is maximized. Here, $P_s(S,a)$ is the probability that sentence $S$ induces sentiment $s$ towards the news-aspect $a$. More specifically, $P(.,.)\in [0,1]$ and $s \in \{positive, negative, neutral\}$. The number of textual modifications to generate $S'$ from $S$ should be minimal such that it appears visually similar to $S$ maintaining grammaticality and overall coherence. However,  in order to modify the sentence, we perform three types of textual transformation operations: \textit{replace, insert} and \textit{delete}. The perturbation process maintains the following conditions: 1) preserves the semantic meaning as much as possible, 2) does not harm fluency and grammatical correctness, and 3) satisfies certain transformation constraints. Note that the probability of each sentiment tag for an aspect in a sentence can be estimated by any pretrained or finetuned classification  model, which takes a sentence and a news-aspect as input and estimates the aspect-specific probability score as output. 



\begin{figure}[htbp]
  \centering
  \includegraphics[width=.47\textwidth]{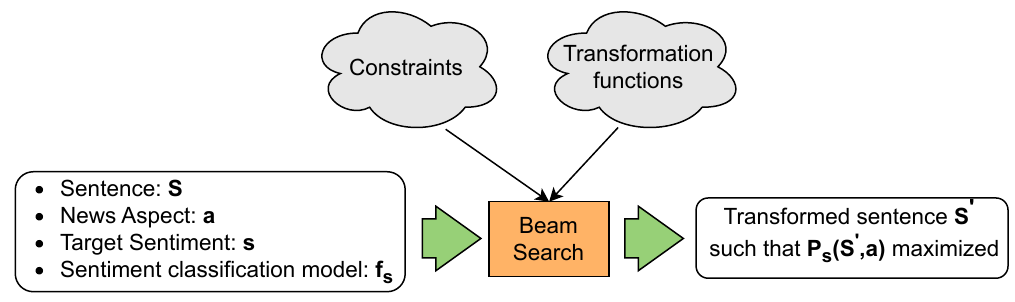}
  \caption{Illustration of Sentiment Polarity Reduction via Contextual Perturbation.}
  \label{fig:attack}
\end{figure}

In the upcoming section, we will comprehensively explain the range of \textit{transformation} operations applied and the \textit{constraints} enforced during contextualized perturbation in sentences. Furthermore, we will illustrate how the proposed system \textit{search} through different transformations to acquire the resulting transformed output (Figure ~\ref{fig:attack}).

\subsection{Contextualized Perturbations}

\label{attack-style}

\mycomment{We have an NLI based sentiment classification model $f$, that takes a sentence - hypothesis pair $<$\textbf{$S$}$,$\textbf{$H_a$}$>$ and outputs probability of each sentiment tag. More specifically, $f_s(S,H) \in [0,1]$ is the probability that the sentence $S$ has a sentiment $s$ towards the aspect $a$ that is mentioned in the hypothesis $H_a$, where $s\in{positive, negative, neutral}$.} 


Textual operations that are applied to the sentences are referenced by $transform()$ function in algorithm \ref{attack_algo}. $transform(S,i)$ denotes the transformation(s) applied on the sentence $S$ at the word index $i$. We have incorporated $3$ types of transformations:

\begin{itemize}
    \item \textbf{replace(S,i):} replace word at index $i$ in sentence $S$ by an alternative word. (e.g. changing ``argued" to ``alleged" in \textit{``He alleged that the CAA ignores Indian Muslims."}).
    This replacement is achieved by replacing $w_i$ with a [MASK] token and predicting a probable token at that position using BERT ~\cite{Bert2018}.

    \item \textbf{insert(S,i):} insert word before the word at index $i$ in sentence $S$ by some word.(e.g. changing \textit{``It hurts ..."} to \textit{``It deeply hurts ..."}. This is achieved by inserting [MASK] token before $w_i$ and predicting a probable token at that position using BERT.

    \item \textbf{delete(S,i):} delete word at index $i$ in sentence $S$. (e.g. changing \textit{``... is a welcome step."} to \textit{``... is a step."}.
\end{itemize}

$transform(S,i)$ may compose of any combination of the above three transformations. e.g. $transform(S,i) = \{replace(S,i), insert(S,i)\}$ is a possible transformation that allows only replacements or insertions at position $i$ in $S$. However, \textit{insert} and \textit{delete} operations helps in generating output sentences of varied length. Since these transformations employ a masked LM that uses context information for predicting the [MASK], we call the transformations contextualized perturbations.

\subsection{Constraints}
We apply a number of constraints in order to ensure that the perturbed sentences are effective in achieving the intended goal of sentiment manipulation while meeting the predefined criteria mentioned in section~\ref{task_def}. The list of all possible constraints that a transformed sentence should satisfy is as follows:
\begin{itemize}
    \item \textbf{RepeatWordModification:} already modified words are not re-modified.
    \item \textbf{StopwordModification:} prevent the modification of stopwords.
    \item \textbf{MaxModificationRate:} allow only modification of a maximum percentage of words in the input sentence, we have performed our experiments fixing this percentage as 10\%.
    \item \textbf{BERTScore:} allow transformations that have at least a minimum cosine similarity between the sentence embeddings obtained using a BERT(~\cite{Bert2018}) model. It is used to maintain the semantics between the input and output sentences~\cite{Zhang2019BERTScoreET}.
    \item \textbf{Entailment:} allow only transformations which generate sentences that entail the input sentence by at least a certain threshold. Entailment is measured using a pretrained MNLI-based RoBERTa model ~\cite{xlm-roberta}. This constraint guarantees that the transformation operations do not alter the core semantic meaning, a crucial requirement within the news domain. 
\end{itemize}

\subsection{Searching through transformations}

We employ \textbf{Beam Search} algorithm to apply transformations and search for the text that maximizes the goal function. We maintain a beam of $k$ current best transformed sentences, then apply the \textit{transform} function at each word index of each of the k sentences and produces new candidate sentences for the next beam. The new beam is constructed by selecting the top $k$ sentences from the candidate sentences. These sentences are sorted based on the higher score of the targeted aspect level sentiment. For the task of identifying the sentiment score related to the given aspect, we leverage an NLI-based RoBERTa model for aspect-specific sentiment classification~\cite{Seoh2021OpenAT} system that has been fine-tuned in our lab. More specifically, $f_s(S,H_a) \in [0,1]$ indicates the probability that the sentence $S$ has a sentiment $s$ towards the aspect $a$ that is mentioned in the hypothesis $H_a$, where $s\in\{positive, negative, neutral\}$.


Algorithm ~\ref{attack_algo} briefly summarizes our algorithm used to perform style transfer.

\RestyleAlgo{ruled}
\begin{algorithm}[!ht]
\caption{Sentiment Transfer via Contextual Perturbation}
\label{attack_algo}
\SetKwData{Left}{left}\SetKwData{This}{this}\SetKwData{Up}{up}
\SetKwFunction{Union}{Union}\SetKwFunction{FindCompress}{FindCompress}
\SetKwInOut{Input}{inputs}\SetKwInOut{Output}{output}\SetKwInOut{Initialize}{initialization}
\SetKwComment{Comment}{/*}{*/}
\Input
{
    $S$: input sentence\;
    $H_a$: hypothesis with aspect {$a$}\;
    $s$: target sentiment\;
    $f$: NLI based sentiment classification model\;
    $k$: Beam width\;
    $C$: list of constraints;
}
\Output
{
    \textbf{$S'$:} modified sentence satisfying $C$ with $f_s(S',H_a)$ maximized
}
\Initialize
{
    $B \gets [S]$ \Comment*[r]{current beam}
    $f_{max} \gets 0$ \Comment*[r]{initialize maximum score for target s}
}
\While{True}{
    $\textbf{N} \gets [\hspace{2pt}]$ \Comment*[r]{candidates for new beam}
    \For{$i\gets1$ \KwTo $len(B)$}{
        \For{$j\gets1$ \KwTo $len(S^i)$} 
        {
            $S^{ij}\gets transform(S^i, j)$ \Comment*[r]{transform the $i^{th}$ sentence in $B$, $S^i$, considering word at index $j$}
            \If{$S^{ij}$ satisfies $C$ and $f_s(S^{ij},H_a) > f_{max}$ }{
                $N.append(S^{ij})$\;
            }
        }
    }
    \If{$N$ is empty}{
        return best solution from $B$\;
    }
    $sort(N)$ in decreasing order of $f$\;
    $B \gets N[1:k]$\;
    $f_{max} \gets f_s(N[1],H_a)$
}

\end{algorithm}

\section{Experiments}
\subsection{Dataset}

For the purpose of bias estimation, we need to extract the aspects present in the news stories and classify the inherent polarity based on the story representation using our NLI based classification models. We also need the aspect labels for each sentence for the purpose of performing style transfer on news text.

We use GDELT~\footnote{\url{https://www.gdeltproject.org/}} (Global Database of Events, Language, and Tone) to locate news articles pertaining to specific subjects. GDELT serves as a publicly accessible repository of global events and activities, complete with links to news articles from diverse newspapers covering significant worldwide occurrences and topics.
Our process involves the aggregation of pertinent news URLs to extract news stories associated with India. Subsequently, we employ a bag-of-words approach, implementing semi-supervised LDA~\cite{6511698}, to identify news articles relevant to the four topics mentioned in Table~\ref{gdelt-data}. To achieve this, we manually annotate approximately 200 documents covering these four subjects, yielding $1353$ annotated sentences. This annotation process is executed using a web-based annotation tool~\cite{https://doi.org/10.48550/arxiv.2108.08184}.

The sentences are annotated with labels specific to both topic-related aspects and aspect-specific sentiments. Each topic boasts a unique set of aspects, with sentiment labels categorized into three types: 1) positive, 2) negative, and 3) neutral. The topics and topic relevant aspects are described in the Appendix \ref{aspect_description_appendix}.

We conduct a comprehensive analysis of the news articles, highlighting significant aspects associated with each news topic. Table~\ref{gdelt-data} offers statistical insights into the extracted data for each topic, while Table~\ref{annotated} outlines the selected aspects and the corresponding number of annotated sentences for each topic.
\begin{table}[htbp]
\centering
\fontsize{5.5}{7}\selectfont
    \begin{tabular}{ccc}
    \hline
    \textbf{\shortstack{News\\Topic}}                                & \textbf{\shortstack{Articles\\ count}} & \textbf{\shortstack{Time\\ range}}     \\ \hline
    Agriculture Act    & 9000                     & 18th Nov, 2019 - 30th Nov, 2021 \\ 
    Citizenship Amendment Bill (CAB)          & 589                      & 9th Sept, 2019 - 25th April, 2020 \\ 
    Demonetization & 3912                     & 8th Sept, 2016 - 30th Nov,2021 \\ 
    COVID-19 pandemic        & 25781                    & 30th Mar, 2020 - 1st Mar, 2022 \\ \hline
    \end{tabular}
    \caption{Topicwise statistics of extracted data}
    \label{gdelt-data}
\end{table}

\begin{table}[htbp]
\center
\fontsize{6}{7}\selectfont
    \begin{tabular}{c|l|c}
    \hline
    \textbf{Topic} & \multicolumn{1}{c|}{\textbf{Aspects}} & \textbf{\shortstack{No of\\ annotated\\ sentences}} \\ \hline
    Agriculture Act & \begin{tabular}[c]{@{}l@{}}farm laws\\ farmer protest\\ government action\\ international involvement\end{tabular} & 495 \\ \hline
    CAB & \begin{tabular}[c]{@{}l@{}}Citizenship Amendment Bill\\ government action\\ protest\\ National Register of Citizens\end{tabular} & 353 \\ \hline
    Demonetization & \begin{tabular}[c]{@{}l@{}}note ban\\ money digitization\\ control black money\\ government action\\ effect on public life\end{tabular} & 253 \\ \hline
    \shortstack{COVID-19\\ pandemic} & \begin{tabular}[c]{@{}l@{}}healthcare situation\\ Lockdown\\ testing and vaccination\\ government action\\ effect on public life\end{tabular} & 252 \\ \hline
    \end{tabular}
    \caption{Topicwise aspects and annotated sentences statistics}
    \label{annotated}
\end{table}


\subsection{Evaluation Metrics}

Evaluating our proposed sentiment transfer model presents a challenge due to the absence of a gold standard or benchmark dataset. However inspired from ~\cite{mir-etal-2019-evaluating} we evaluate the performance based on four factors: 1) change in neutrality, 2) fluency, 3) content preservation and 4) Levenshtein distance.

\begin{itemize}
    \item \textbf{Change in neutrality}: We evaluate the performance of our style transfer framework using the change in the prediction of target sentiment probability. Henceforth we have used the term \textit{neutrality} to denote the neutral probability prediction. In most of the experiments, we have used the change in \textit{neutrality} as the performance metric. When we used \textit{positive} and \textit{negative} as target sentiments, we measured change in \textit{positivity} and \textit{negativity} as the performance metric.
    \item \textbf{Fluency}: The fluency of the sentences is measured by \textit{perplexity}. We employ GPT2~\cite{Radford2019LanguageMA} to calculate the perplexity of the sentence.
    
    \item \textbf{Content Preservation}: We measure the correctness of the transformed sentences by measuring to what extent the output sentence entails the original input sentences using a pre-trained RoBERTa MNLI model ~\cite{xlm-roberta}.
    
    \item \textbf{Levenshtein distance}: We also measure the similarity between the input and output sentences using 1 - Normalized Levenshtein distance ~\cite{normalized_dist} as the metric.
\end{itemize}

\subsection{Result Analysis}

\subsubsection{Comparision of Transformation methods}
\label{transform-exp}

In Table~\ref{table:style}, we evaluate the performance of our adversarial attack-based sentiment transfer framework using different transformation methods. Notably, higher beam sizes consistently improve performance across all methods.

\begin{table}[htbp]
\center
\fontsize{7.5}{8}\selectfont
\setlength\tabcolsep{3pt}
\begin{tabular}{l|c|c|c|c}
\hline
\begin{tabular}[c]{@{}l@{}}\textbf{Transformation} \\ \end{tabular} & \begin{tabular}[c]{@{}l@{}}\textbf{Beam}\\ \textbf{width}\end{tabular} & \begin{tabular}[c]{@{}l@{}}\textbf{Average} \\ \textbf{entailment}\end{tabular} & \begin{tabular}[c]{@{}l@{}}\textbf{Average} \\ \textbf{Levenshtein} \\ \textbf{similarity}\end{tabular} & \begin{tabular}[c]{@{}l@{}}\textbf{Average} \\ \textbf{neutrality} \\ \textbf{change}\end{tabular} \\ \hline
Replace & 1 & 0.55 & 0.86 & 0.26 \\ \hline
Replace & 2 & 0.57 & 0.85 & 0.28 \\ \hline
Replace & 3 & 0.58 & 0.85 & 0.29 \\ \hline
Delete & 1 & \textbf{0.61} & 0.69 & 0.24 \\ \hline
Delete & 2 & 0.58 & 0.67 & 0.25 \\ \hline
Delete & 3 & 0.57 & 0.66 & 0.26 \\ \hline
Insert & 1 & 0.47 & 0.89 & 0.10 \\ \hline
Insert & 2 & 0.46 & 0.89 & 0.11 \\ \hline
Insert & 3 & 0.45 & \textbf{0.90} & 0.12 \\ \hline
Replace+Insert & 1 & 0.51 & 0.88 & 0.26 \\ \hline
Replace+Insert & 2 & 0.55 & 0.88 & 0.29 \\ \hline
Replace+Insert & 3 & 0.53 & 0.87 & \textbf{0.30} \\ \hline
Replace+Delete & 3 & 0.58 & 0.86 & 0.29 \\ \hline
\end{tabular}
\caption{Performance comparison for various search configurations for style transfer - BERTScore threshold was set to 0.95 and maximum 10\% word perturbations were allowed. Goal was to maximize the probability of neutral sentiment}
\label{table:style}
\end{table}

We observe that using the \textit{insertion} method alone results in limited neutrality change and lower average entailment scores compared to other methods. Specifically, \textit{insertion} achieves an average neutrality change of around $0.1$, while other methods surpass $0.2$. Both \textit{deletion} and \textit{replacement} methods perform similarly in terms of neutrality change, although \textit{deletion} exhibits a decrease in textual similarity.

Interestingly, combining \textit{replacement} and \textit{insertion} as transformation methods does not significantly improve neutrality change compared to using \textit{replacement} alone. For instance, \textit{replacement} with a beam size of 3 results in a neutrality change of $0.28$, while \textit{replacement + insertion} with a beam width of 3 achieves a similar value of $0.29$. However, it is important to note that using \textit{replacement + insertion} doubles the time required for style transfer compared to using \textit{replacement} exclusively.

In summary, our findings suggest that relying solely on the \textit{replacement} method for text transformation can achieve satisfactory neutrality change with favorable entailment values, as compared to other transformation combinations, given that \textit{deletions} and \textit{insertions} may remove or introduce additional information into the sentences.

\subsubsection{Effect of beam size in Performance}
\label{beam-size-exp}

\begin{figure}[htbp]
    \centering
    \includegraphics[width=0.85\linewidth]{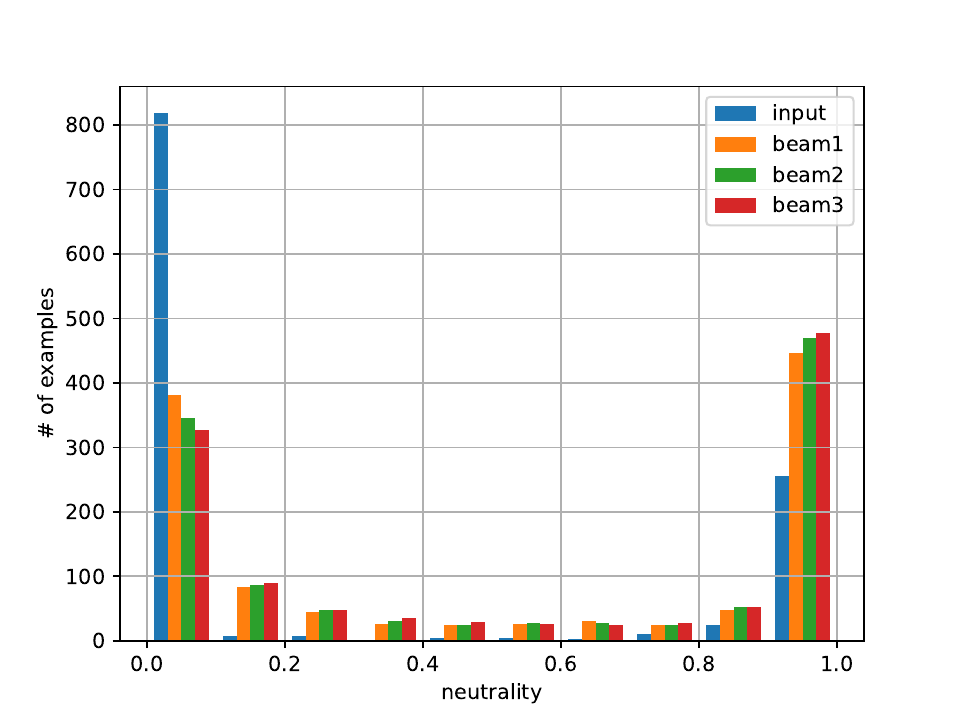}
    \caption{Number of examples for corresponding neutrality change for various beam widths. Goal: to maximize output neutrality, Transformation: only replacement, min BERTScore: 0.95.}
    \label{fig:beam_change}
\end{figure}

Figure~\ref{fig:beam_change} shows a histogram of the number of examples with their neutrality values corresponding to each 0.1-sized bin varying from 0 to 1 on the x-axis. The blue bar represents the majority of pairs with neutrality scores between 0 and 0.1, mainly comprising sentences with positive or negative sentiments. The other bars display histograms for three settings, each with different beam widths (1, 2, and 3). In all three settings, the target sentiment was "neutral," and the transformation method was word replacement with a minimum BERTScore constraint of 0.95. It's noteworthy that as the beam width increases, the number of examples with high neutrality scores (0.9-1.0) also rises. This trend confirms that a larger beam width leads to better style transfer performance, reducing the chances of suboptimal results.

\subsubsection{Effect of length of the sentence on Entailment}

\label{length-exp}

\begin{figure}[htbp]
    \centering
    \subfloat[x-axis: length of sentence, y-axis: mean neutrality change] {{\includegraphics[width=0.5\linewidth]{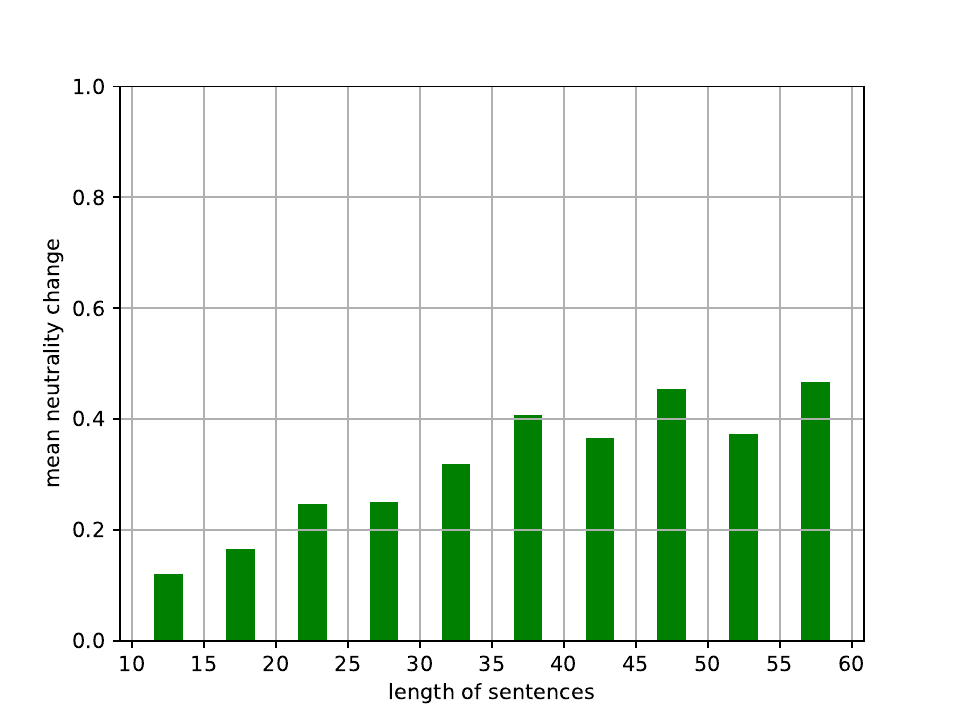}}}
    \hfill
    \subfloat[x-axis: length of sentence, y-axis: mean entailment] {{\includegraphics[width=0.5\linewidth]{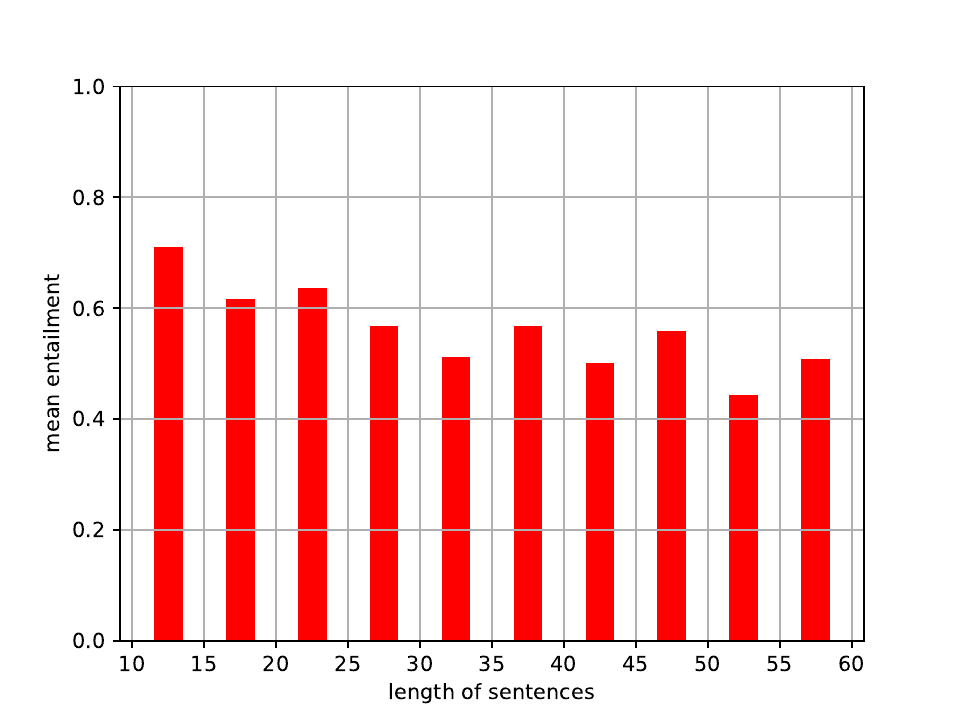}}}
    \caption{Variation of neutrality change and entailment with length of sentences. Goal: to maximize output neutrality, Transformation: only replacement.}
    \label{fig:length_change}
    \setlength{\belowcaptionskip}{10pt}
\end{figure}

Figure~\ref{fig:length_change}(a) shows the mean neutrality change for various lengths of sentences. The lengths of the sentences are grouped into bin sizes of 5 in the plot. Out of the $1353$ examples, 1070 sentences had lengths between 10 to 60, hence the x-axis varies in that range. The transformation method used in the experiment was word-replacement with the minimum BERTScore constraint of 0.95. Search was done using a beam width of 3.

We observe that as the length of the sentence increases, the neutrality change increases. This is mainly because in longer sentences, we can perform modifications to more number of words. But the higher neutrality change in longer sentences comes at a cost of decrease in entailment as well, as evident from figure~\ref{fig:length_change}(b). Since more modifications are possible and we use a pretrained model for measuring entailment and in longer sentences it has to capture larger context, this may lead to loss in entailment with the input sentence. Detailed discussions on the variations in model performance under different experimental setups can be found in the appendix section.

\subsubsection{Comparison with ChatGPT}

\label{gpt-exp}

We try to perform the task of sentiment transfer using ChatGPT ~\footnote{\url{https://chat.openai.com/}} based on large language model (LLM) GPT-3.5. We use the ChatGPT API to use its pretrained model using the following prompt as input:

\textit{
Given
Input sentence: "$<$sentence$>$",
Aspect of the sentence: "$<$aspect$>$",
Aspect description: "$<$aspect description$>$",
Modify the input sentence minimally and try not to add the aspect description in the output sentence, while preserving its meaning such that the sentiment towards the aspect "$<$aspect$>$" is as neutral as possible.
Please return only the output sentence.
}

where $<...>$ indicates a placeholder. Note we have also added an aspect decription in the prompt, since ChatGPT has never seen these aspects before, but expect it to perform reasonably since it is an LLM trained on huge variety of texts.

\begin{table}[htbp]
\center
\fontsize{6}{8}\selectfont
\setlength\tabcolsep{3pt}
\begin{tabular}{l|l|l|l}
\hline
\textbf{Style Transfer method} & \textbf{\begin{tabular}[c]{@{}l@{}}Average \\ entailment\end{tabular}} & \textbf{\begin{tabular}[c]{@{}l@{}}Average \\ Levenshtein \\ similarity\end{tabular}} & \textbf{\begin{tabular}[c]{@{}l@{}}Average \\ neutrality \\ change\end{tabular}} \\ \hline 
ChatGPT & 0.90 & 0.52 & 0.13 \\ \hline \hline
\begin{tabular}[c]{@{}l@{}}Our method with beam width 3, \\ using only replacement \\ transformation and minimum \\ BERTScore 0.95\end{tabular} & 0.58 & 0.85 & 0.29 \\ \hline
\end{tabular}
\caption{Comparison between ChatGPT and our method for sentiment transfer}
\label{table:gpt}
\end{table}

From table~\ref{table:gpt}, we can observe that in terms of neutrality change our method outperforms ChatGPT, but on the other hand the mean entailment of the output sentences generated by ChatGPT is much higher (0.9) as compared to our method (0.58). Also, the Levenshtein similarity is less in case of ChatGPT (0.52), since it changes the structure of the whole input sentence, instead of just modifying a few words. Better prompt design can be used to leverage the power of ChatGPT in style transfer in a finer way.

\subsubsection{Human Evaluation}

We also perform human evaluation of the outputs generated by our sentiment transformation framework. For this assessment, we randomly selected $30$ sentences from the evaluation set and their rephrased counterparts produced by our proposed methods. These sentence pairs were evaluated by a panel of three raters, which included a Ph.D. scholar specializing in English literature and humanities, as well as two Ph.D. students with a background in computer science. The raters were instructed to score each rephrased sentence based on three criteria: content preservation, neutrality, and fluency, using a scale from 1 to 5 (with 5 indicating the highest score). The criteria encompassed:

\begin{itemize}
    \item Content Preservation: Assessing the extent to which the rephrased sentence preserved the original meaning and context.
    
    \item Neutrality: Evaluating the effectiveness of the rephrased sentence in reducing sentiment intensity, aligning it with the goal of neutrality.

    \item Fluency: Focusing on the grammatical correctness and overall coherence of the rephrased sentence.
\end{itemize}

The averaged findings across raters for different transformation settings are reported in Table \ref{table:human_eval}. Across all three criteria, the mean Pearson correlation coefficient for the scores assigned by the three evaluators exceeds $0.67$, affirming substantial inter-evaluator agreement.

\begin{table}[htbp]
\center
\fontsize{9}{10}\selectfont
\setlength\tabcolsep{3pt}

\begin{tabular}{c|c|c|c}
\hline
\begin{tabular}[c]{@{}c@{}}Transformation \\ Model\end{tabular} & \begin{tabular}[c]{@{}c@{}}Content \\ Preservation\end{tabular} & Neutrality & Fluency \\ \hline
Only Replace                                                    & 3.21                                                            & 3.53       & 3.9     \\ \hline
Only Delete                                                     & 2.81                                                            & 3.11       & 2.98    \\ \hline
Only Insert                                                     & 2.53                                                            & 2.89       & 3.17    \\ \hline \hline
ChatGPT output                                                  & 3.98                                                            & 4.03       & 4.4     \\ \hline
\end{tabular}

\caption{Human Assessment of Results: Evaluating Outputs from Adversarial Attack-Based Transformations (Beam Size 3) and ChatGPT Models, Using Three Criteria: content, neutrality (sentiment) and fluency}
\label{table:human_eval}
\end{table}

\section{Related Works}

Sentiment transfer is a growing area of research in natural language processing (NLP) and sentiment analysis. The task of sentiment transfer involves altering the sentiment of a given text while preserving its content and meaning. Several research papers have explored different approaches and techniques for sentiment transfer~\cite{li-etal-2018-delete, Xu2018UnpairedST, lample-etal-2018-phrase, Krishna2020ReformulatingUS}. The existing approaches for transferring sentiment revolve around back-translation~\cite{Lample2018MultipleAttributeTR, Xu2019PrivacyAwareTR}, Variational Auto-encoder~\cite{duan-etal-2020-pre}, encoder-decoder with discriminator~\cite{Majumder2021UnsupervisedEO, Romanov2018AdversarialDO}, pretraining~\cite{Zhou2021ImprovingSP} etc. However, there are few works on the fine-grained transfer of sentiments where the objective is to revise the sentence to change the intensity of the sentiment. The work by~\cite{Liao2018QuaSESE} involves utilizing a model based on Variational Autoencoder (VAE) and training it with pseudo-parallel data to enable sentence editing for fine-grained sentiment transfer. ~\cite{Luo2019TowardsFT} design a cycle reinforcement learning algorithm for manipulating the fine-grained sentiment intensity of the output sentence. ~\cite{Liu2021PoliticalDO} exploit attribute-aware word embeddings for the identification of political bias in news articles and propose a probabilistic algorithm for depolarizing the text. ~\cite{Liu2021MitigatingPB} present  a Reinforcement learning framework for mitigating political bias in news text. All these works manipulate the sentiment or sentiment polarity of the overall sentences. A few works change the aspect-specific sentiment labels in sentences~\cite{narayanan-sundararaman-etal-2020-unsupervised}. In contrast we aim to rewrite the sentence to manipulate the sentiment intensity of the targeted aspect present in the sentence.

\section{Conclusion}

Our research aims to reduce sentiment polarity in news reporting using advanced techniques like adversarial attack-based perturbations and large language models such as ChatGPT with prompt engineering, striving for more balanced and impartial news narratives. Looking forward, we are exploring avenues for improving efficiency in news sentiment mitigation, including automating the identification of news aspects within sentences. We also aim to address implicit sentiment and expand our techniques to cover a wider range of languages and domains, particularly in low-resource language contexts, with the ultimate goal of advancing news reporting towards greater equity and impartiality.



\bibliography{acl_latex}

\newpage
\onecolumn
\appendix
\twocolumn
\section{Appendix}
\label{sec:appendix}

\subsection{Topic specific Aspect Descriptions}
\label{aspect_description_appendix}
\begin{itemize}
    \item Agriculture Act~\footnote{\url{https://en.wikipedia.org/wiki/2020_Indian_agriculture_acts}}:
    \begin{itemize}
        \item government action:The action made by government, PM, Agriculture minister, and the police. These actions include planning for the meeting, requesting a meeting, police firing, lathi-charge, arrest made by police etc.
        \item International involvement:Protest against farm law outside India (UK, USA, Canada, etc), discussion on governmentt action in foreign countries.
        \item Farm law:reports on farm laws or farm bills or Agriculture Acts, its long term effect on farmers, economy.
        \item farmer protest:Reports depicting any type of farmer protest against the Farm Bills such as rally, road blockage, tractor rally, destruction of public property, etc.
    \end{itemize}

    \item Demonetization~\footnote{\url{https://en.wikipedia.org/wiki/2016_Indian_banknote_demonetisation}}:
        \begin{itemize}
            \item effect on public life:Information related to problems faced by the public, money shortage, unemployment (demonetization’s effect on public life).
            \item government action:The action made by government, Prime minister, Finance Minister E.g. meeting, discussion on Note Ban.
            \item control black money:black money reduction or recovery after demonetization.
            \item note ban:note ban and effects of demonization.
            \item money digitization:digitization of currency
        \end{itemize}

    \item CAB~\footnote{\url{https://en.wikipedia.org/wiki/Citizenship_(Amendment)_Act,_2019}}:
        \begin{itemize}
            \item government action:The action made by government, PM, HM, police, Ruling Party. These actions include discussion in Parliament, police firing, lathi-charge, arrest against protesters etc.
            \item citizenship amendment bill:reports on citizenship amendment bill (CAB) and its effects.
            \item protest:protest against the bill. Eg. protest in Assam, shaheen bagh protest, student protest in the university.
            \item National Register of Citizens (NRC):National Register of Citizens (NRC)
        \end{itemize}

    \item COVID-19 Control~\footnote{\url{https://en.wikipedia.org/wiki/COVID-19_pandemic_in_India}}:
        \begin{itemize}
            \item effect on public life:Displacement of migrant workers, effect on students, jobless and unemployment.
            \item healthcare situation:Shortage of healthcare, scarcity of oxygen, hospital beds, health personnel infected, etc.
            \item Lockdown: Imposing lockdown in the whole country and states.
            \item testing and vaccination: Testing of covid virus transmission and vaccine production and distribution.
            \item government action:Relief and welfare initiatives made by the government, meetings held by the government, actions taken by the police, etc.

        \end{itemize}

\end{itemize}

\subsection{Effect of beam size in Performance}
\label{beam-size-exp_appendix}

\begin{figure}[!ht]
    \centering
    \subfloat[Beam width = 1] {{\includegraphics[width=0.8\linewidth]{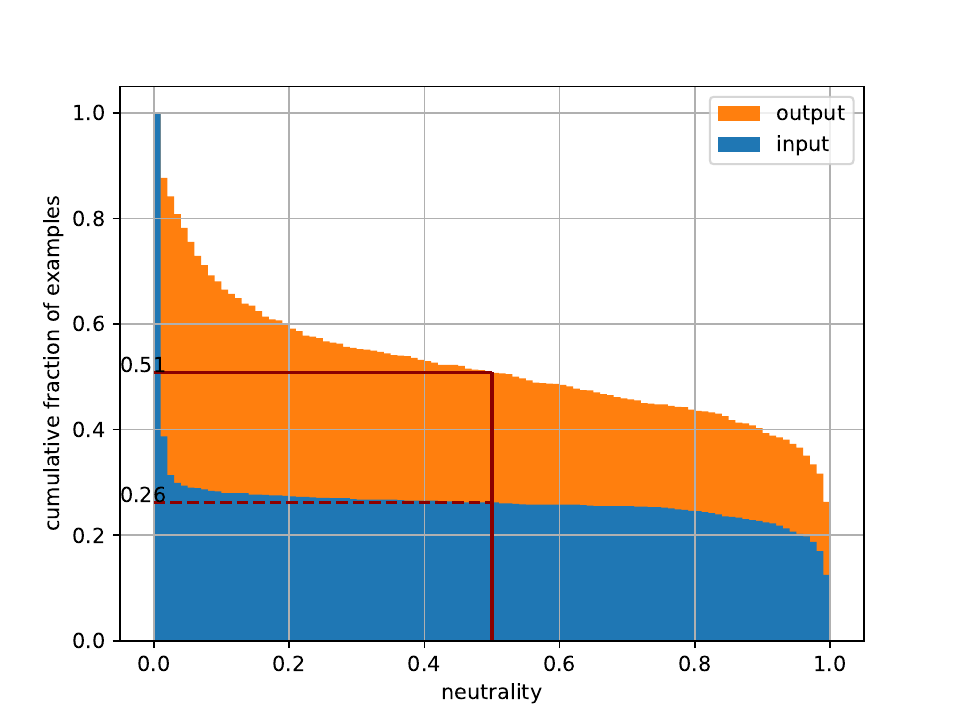}}}
    \hfill
    \subfloat[Beam width = 3] {{\includegraphics[width=0.8\linewidth]{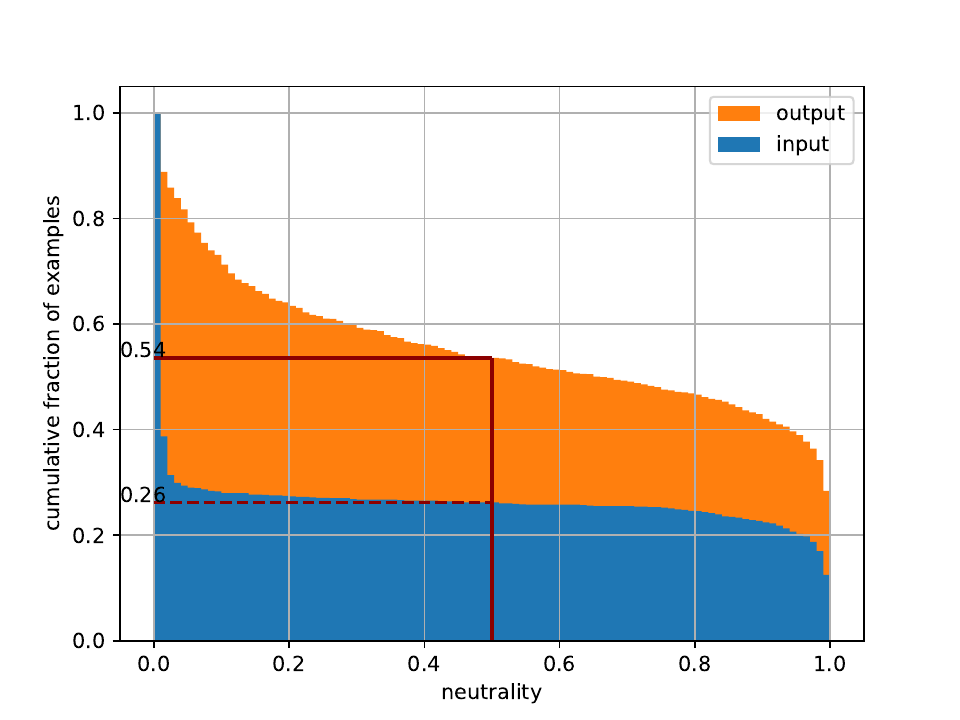}}}
    \caption{Cumulative fraction of examples vs neutrality. Goal: to maximize output neutrality, Transformation: only replacement, min BERTScore:0.95}
    \label{fig:beam_cumulative}
    \setlength{\belowcaptionskip}{10pt}
\end{figure}

Figures ~\ref{fig:beam_cumulative}(a) and (b) show the cumulative fraction of examples with their neutrality values as predicted by our NLI model greater than certain value. Here the neutrality values are binned into 100 bins of size 0.01 each. Assuming that an instance is classified as neutral if the probability of neutral classification is more than the sum of probabilities of positive and negative classification, we choose a marker at neutrality 0.5 to detect the fraction of examples that are classified as neutral using our NLI model. In the case when beam width was taken to be 1, 51\% of the examples have a neutrality score $>$ 0.5, while in case of beam width 3, the percentage improved to 54\%. Thus, there was a 3\% increase in the number of transformed sentences that were classified as neutral.

\begin{figure}[htbp]
    \centering
    \includegraphics[width=0.9\linewidth]{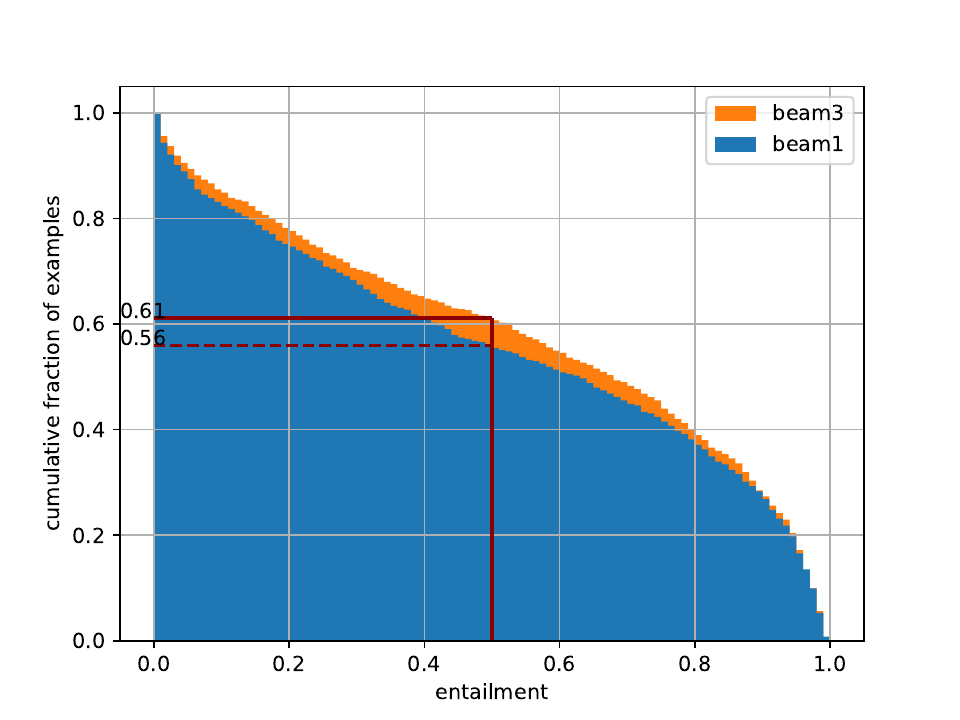}
    \caption{Cumulative fraction of examples vs entailment. Goal: to maximize output neutrality, Transformation: only replacement, min BERTScore:0.95}
    \label{fig:beam_entail_change}
\end{figure}

Figure~\ref{fig:beam_entail_change} shows that increasing the beam size also leads to a better entailment of the transformed sentences with the input sentences. In case of beam width 3, 61\% of the examples have an entailment of $>$ 0.5 with the input sentences which is 5\% more than in the case of beam width 1.

\subsection{Entailment variation with neutrality of input sentences}
\mycomment{
\begin{figure}[!ht]
    \centering
    \subfloat[Mean neutrality change vs input neutrality] {{\includegraphics[width=0.35\textwidth]{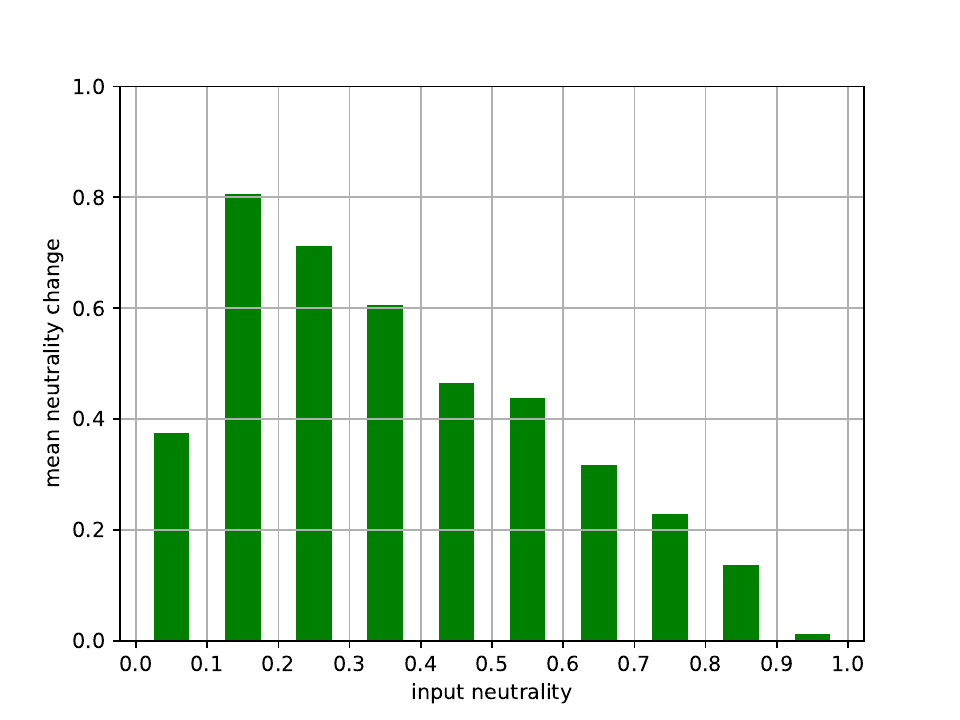}}}
    \hfill
    \subfloat[Mean output neutrality vs input neutrality] {{\includegraphics[width=0.35\textwidth]{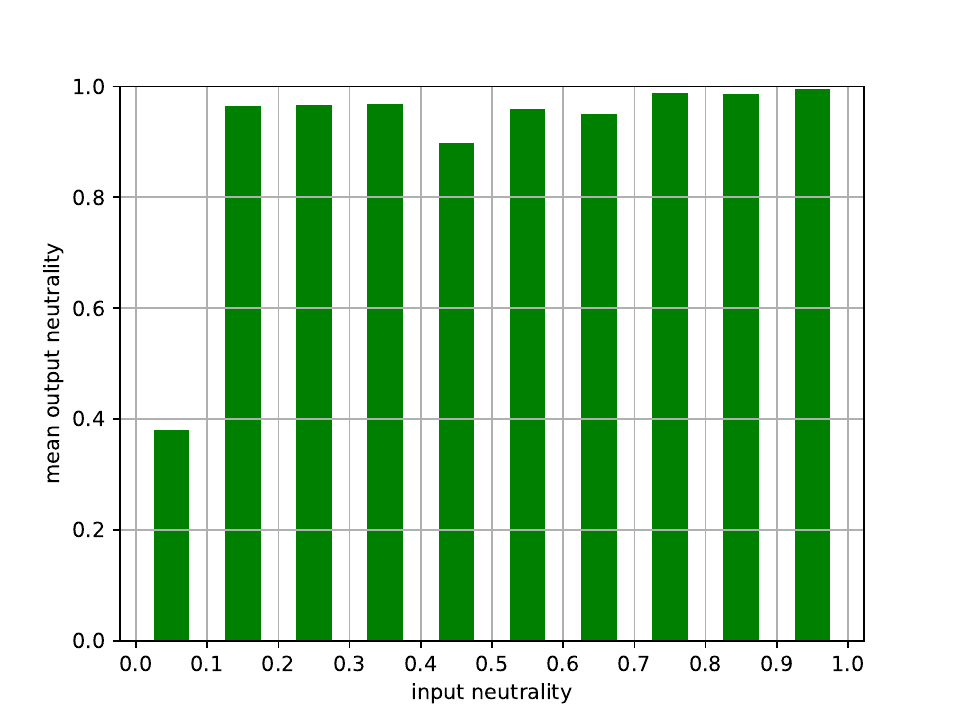}}}
    \hfill
    \subfloat[Heatmap of neutrality change vs input neutrality] {{\includegraphics[width=0.35\textwidth]{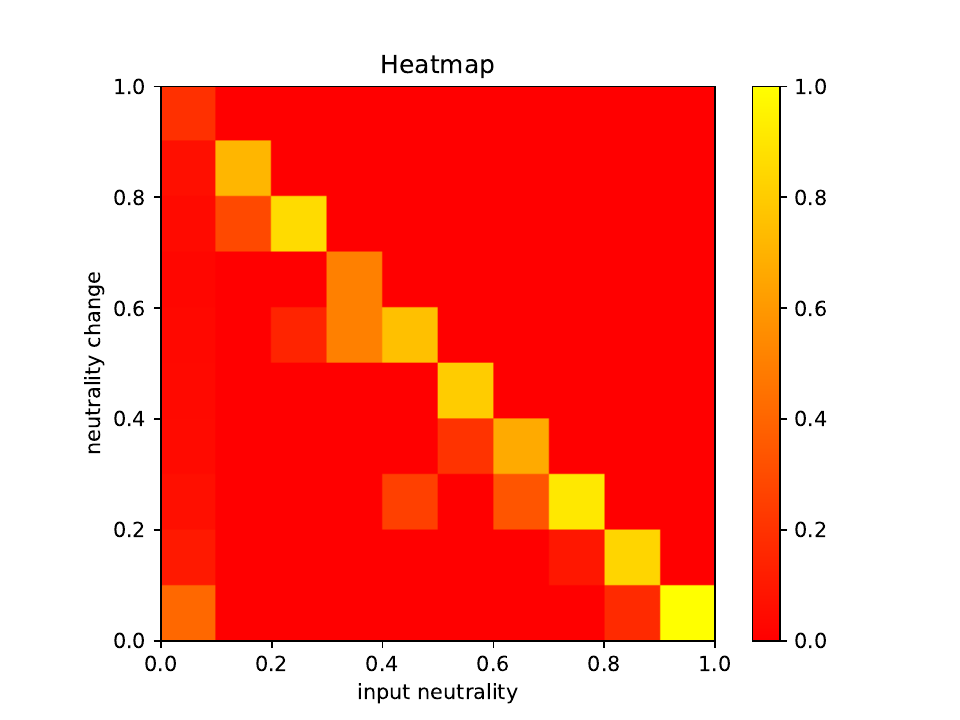}}}
    \caption{Variation of neutrality change vs input neutrality. Goal: to maximize output neutrality, Transformation: only replacement, min BERTScore:0.95, Beam width:3}
    \label{fig:neu_neu}
    \setlength{\belowcaptionskip}{10pt}
\end{figure}

Figure~\ref{fig:neu_neu}(a) shows the mean neutrality change corresponding to the input neutrality values which vary in step sizes of 0.1. As evident from the figure, the sentences which were less neutral had the greatest neutrality changes and the neutrality changes decrease with increase in input neutrality. This is because not more changes are required to make already neutral sentences more neutral. 

We also observe that the least neutral or most polarized (neutrality in [0-0.1]), showed lower increase in neutrality as compared to more neutral input sentences. This is because of incorporating constraints such as BERTScore to maintain semantic similarity. From figure~\ref{fig:neu_neu}(b) we can observe that more neutral sentences tend to be modified to sentences with high neutrality values.

Figure~\ref{fig:neu_neu}(c) shows a heatmap where the x axis shows input neutrality values ranging from 0 to 1 in step size of 0.1 and the y axis shows the neutrality change values ranging from 0 to 1 in step size of 0.1. It appears as if sentences with input neutrality above 0.1 were modified such that the sum of input neutrality and neutrality change is close to 1.0.
}

\begin{figure}[htbp]
    \centering
    \subfloat[Mean entailment vs neutrality change] {{\includegraphics[width=0.35\textwidth]{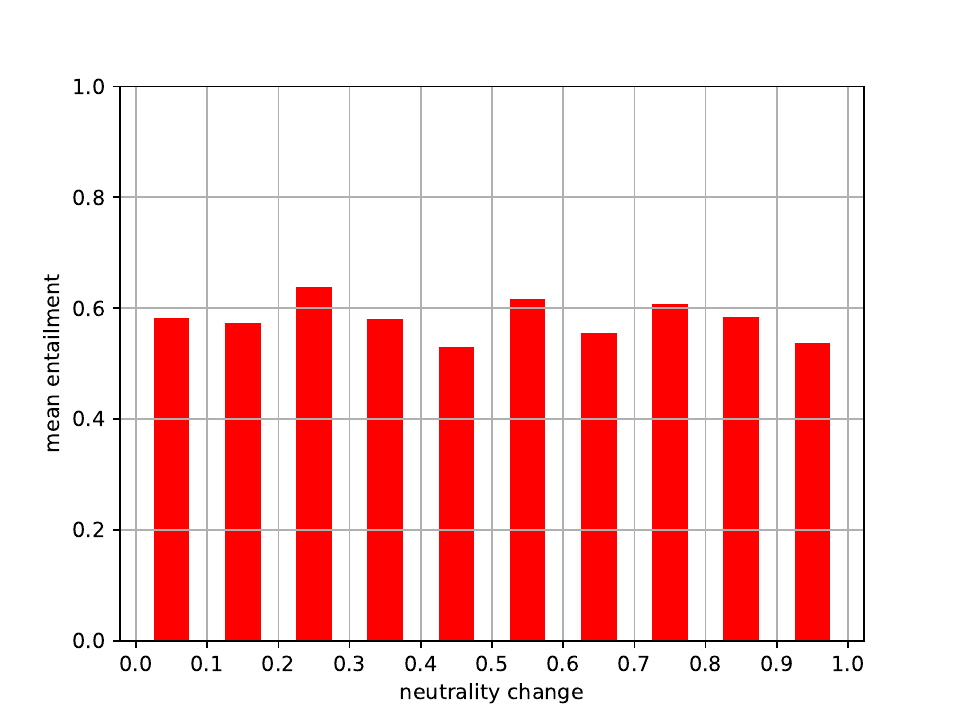}}}
    \hfill
    \subfloat[Heatmap of entailment vs neutrality change] {{\includegraphics[width=0.35\textwidth]{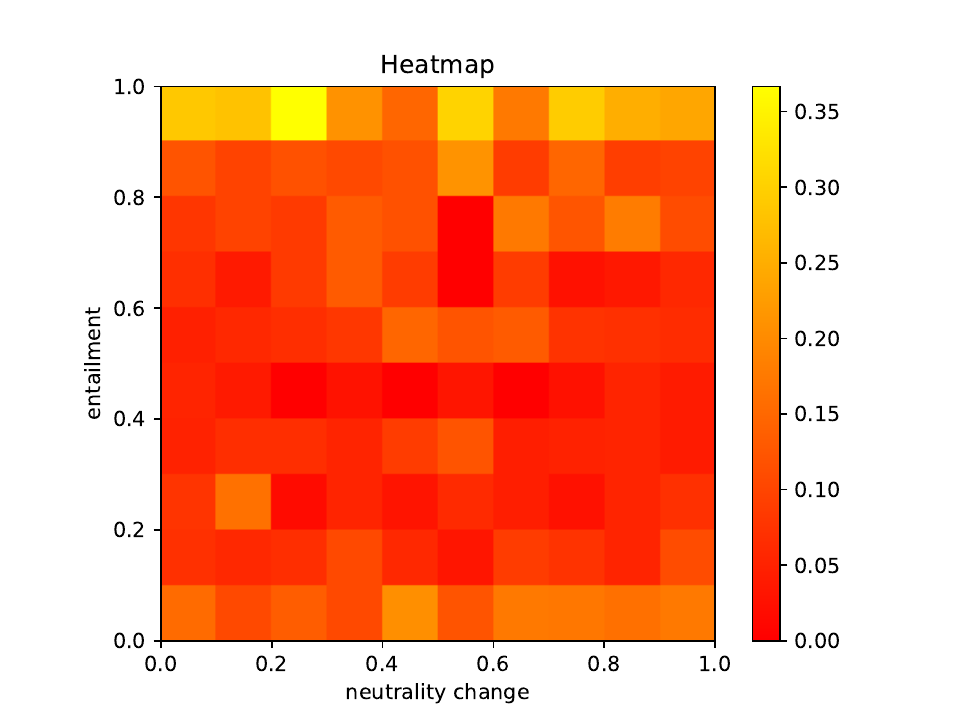}}}
    \caption{Variation of entailment with neutrality change. Goal: to maximize output neutrality, Transformation: only replacement, min BERTScore:0.95, Beam width:3}
    \label{fig:neu_ent}
    \setlength{\belowcaptionskip}{10pt}
\end{figure}

Figure~\ref{fig:neu_ent}(a) and (b) show how the entailment varies with the neutrality change of the input sentences.We see that the mean entailment has a very small variation with neutrality change. It implies that higher neutrality change does not come at a cost of loss in entailment of the modified sentence. The heatmap shows that in most of the cases the output sentence had a high entailment with the original one.

\subsection{Effect of BERTScore (Cosine similarity) as a constraint}

\label{bert-score-exp}

\begin{table}[htbp]
\center
\fontsize{6}{7}\selectfont
\setlength\tabcolsep{3pt}
\begin{tabular}{l|l|l|l}
\hline
\textbf{\begin{tabular}[c]{@{}l@{}}Minimum \\ Cosine \\ Similarity \end{tabular}} & \textbf{\begin{tabular}[c]{@{}l@{}}Average \\ entailment\end{tabular}} & \textbf{\begin{tabular}[c]{@{}l@{}}Average \\ Levenshtein \\ similarity\end{tabular}} & \textbf{\begin{tabular}[c]{@{}l@{}}Average \\ neutrality \\ change\end{tabular}} \\ \hline
0.85 & 0.39 & 0.85 & 0.39 \\ \hline
0.9 & 0.45 & 0.84 & 0.37 \\ \hline
0.95 & 0.58 & 0.85 & 0.27 \\ \hline
\end{tabular}
\caption{Performance comparision for different values of minimum BERTScore (cosine similarity). In all 3 cases, maximum 10\% word perturbations were allowed. Search had a beam width of 3 and only transformation used was replacement.}
\label{table:cosine}
\end{table}

Table~\ref{table:cosine} compares the performance of style transfer under the variation of the minimum BERTScore (cosine similarity) constraint. We observe that though decreasing the minimum cosine similarity constraint gives better neutrality change in the output, it comes at a cost of loss in entailment between the input and output sentences. This difference in average entailment is large between minimum BERTScore of 0.85 and 0.95. The levenshtein similarity almost remain unvaried.

\begin{figure}[htbp]
    \centering
    \includegraphics[width=0.9\linewidth]{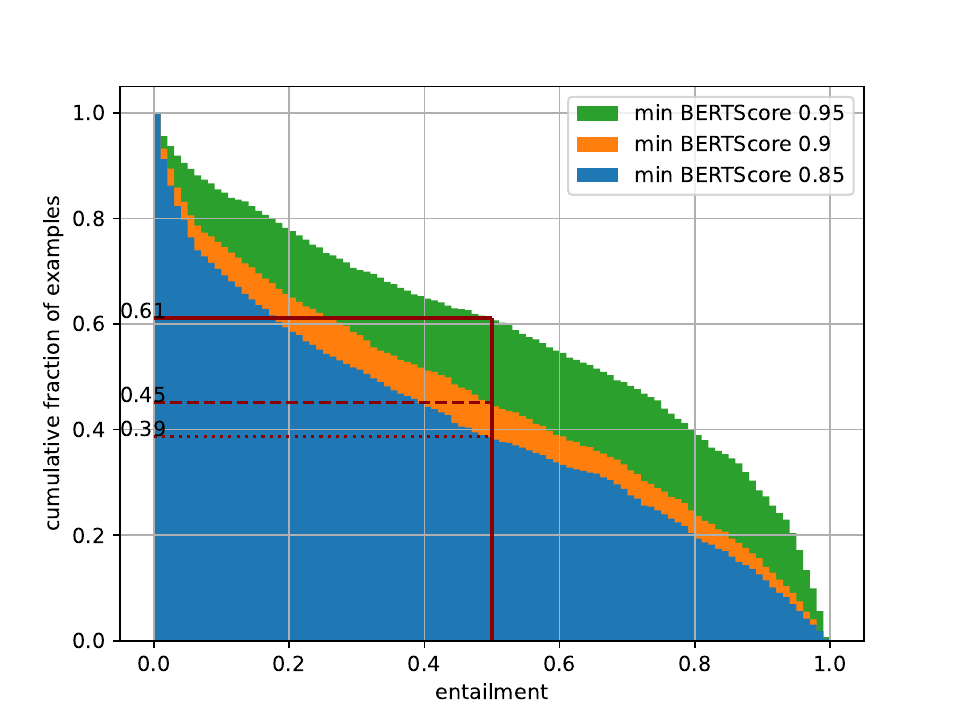}
    \caption{Cumulative fraction of examples vs entailment. Goal: to maximize output neutrality, Transformation: only replacement, Beam width:3}
    \label{fig:cos_entail_change}
\end{figure}

Figure~\ref{fig:cos_entail_change} supports the results in table~\ref{table:cosine}. It shows the cumulative fraction of examples with their entailment greater than a certain value. Here the entailment values are binned into 100 bins of size 0.01 each. While 61\% of the transformed sentences have more than 0.5 entailment with the original text in case of 0.95 minimum BERTScore, only 39\% of the output sentences have more than 0.5 entailment in case of 0.85 minimum BERTScore. This leads to the conclusion that BERTScore is an important constraint for preserving the semantics of the input sentence in the transformed sentence.

Table~\ref{cos_example} shows how just changing the minimum BERTScore from 0.95 to 0.9 can change the meaning of the transformed sentence drastically. With minimum BERTScore of 0.9, the modified sentence tells that the decision was to declare Rs 500 and Rs 1000 notes ``valid", which implies that the semantics of the original sentence is not preserved in the output.

\begin{table*}[htbp]
\resizebox{\textwidth}{!}{%
\begin{tabular}{|l|l|l|l|l|l|}
\hline
\textbf{Input sentence} & \textbf{\begin{tabular}[c]{@{}l@{}}Original \\ sentiment\end{tabular}} & \textbf{\begin{tabular}[c]{@{}l@{}}Minimum\\ BERTScore\end{tabular}} & \textbf{Modified sentence} & \textbf{\begin{tabular}[c]{@{}l@{}}Neutrality \\ change\end{tabular}} & \textbf{\begin{tabular}[c]{@{}l@{}}Entailment \\ with input\end{tabular}} \\ \hline
\multirow{2}{*}{\begin{tabular}[c]{@{}l@{}}The decision to declare Rs 500 \\ and Rs 1000 notes invalid from \\ this midnight is a welcome move.\end{tabular}} & \multirow{2}{*}{positive} & 0.9 & \begin{tabular}[c]{@{}l@{}}The decision to declare Rs 500 \\ and Rs 1000 notes \colorbox{red}{valid} from this \\ midnight is a \colorbox{yellow}{significant} \colorbox{yellow}{decision}.\end{tabular} & 0.80 & 0.38 \\ \cline{3-6} 
 &  & 0.95 & \begin{tabular}[c]{@{}l@{}}The decision to declare Rs 500 and \\ Rs 1000 \colorbox{yellow}{denominations} invalid from \\ this midnight is a \colorbox{yellow}{significant} move.\end{tabular} & 0.58 & 0.88 \\ \hline
\end{tabular}%
}
\caption{An example showing the effect of minimum BERTScore constraint on the transformed sentence}
\label{cos_example}
\end{table*}

\subsection{Effect of Entailment as a constraint}

\label{ent-exp}

\begin{table}[htbp]
\center
\fontsize{6}{7}\selectfont
\setlength\tabcolsep{3pt}
\begin{tabular}{l|l|l|l}
\hline
\textbf{\begin{tabular}[c]{@{}l@{}}Minimum \\ entailment\end{tabular}} & \textbf{\begin{tabular}[c]{@{}l@{}}Average \\ entailment\end{tabular}} & \textbf{\begin{tabular}[c]{@{}l@{}}Average \\ Levenshtein \\ similarity\end{tabular}} & \textbf{\begin{tabular}[c]{@{}l@{}}Average \\ neutrality \\ change\end{tabular}} \\ \hline
0.3 & 0.61 & 0.86 & 0.20 \\ \hline
0.4 & 0.63 & 0.87 & 0.18 \\ \hline
\end{tabular}
\caption{Performance of style transfer adding minimum entailment as constraint. Goal was to maximize output neutrality.}
\label{ent-ent}
\end{table}

Table~\ref{ent-ent} shows how the style transfer task performs if we add an additional constraint that the output sentence should have a minimum entailment with the input sentence. The transformation method used was replacement only with constraints of minimum 0.95 BERTScore and minimum 0.3 or 0.4 entailment. The search was done using a beam width of 3. As compared to the results in first three rows of table~\ref{table:style}, we can observe that the average entailment shows an overall improvement and so does the Levenshtein similarity. But the gain in entailment leads to a decrease in the average neutrality change. Since the search is now more constrained, it is difficult to obtain more neutral sentences.

\begin{figure}[htbp]
    \centering
    \includegraphics[width=0.9\linewidth]{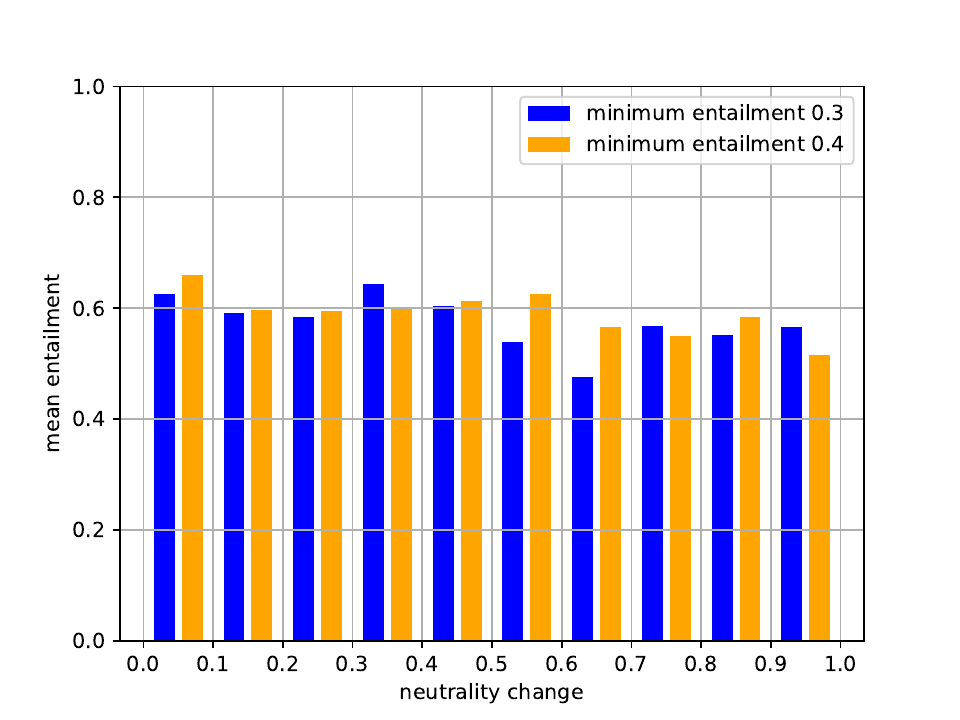}
    \caption{Mean entailment vs neutrality change. Goal: to maximize output neutrality, Transformation: only replacement, min BERTScore:0.95, Beam width:3}
    \label{fig:ent_neu_ent}
\end{figure}

In figure ~\ref{fig:neu_ent}(a), we saw how the entailment did not follow any increasing or decreasing pattern with increasing neutrality change. But from figure~\ref{fig:ent_neu_ent}, we can see there is slight decrease in mean entailment as the neutrality change increases, though the decrease is not steady. The sentences with neutrality change of [0.9-1.0] have the least mean entailment. This reflects that adding minimum entailment requirement as a constraint, leads to better search of neutral sentences.

\subsection{Style transfer for positive and negative sentiments}

\label{pos-neg-exp}

\begin{table}[htbp]
\center
\fontsize{6}{7}\selectfont
\setlength\tabcolsep{3pt}
\begin{tabular}{l|l|l|l}
\hline
\textbf{\begin{tabular}[c]{@{}l@{}}Target \\ sentiment\end{tabular}} & \textbf{\begin{tabular}[c]{@{}l@{}}Average \\ entailment\end{tabular}} & \textbf{\begin{tabular}[c]{@{}l@{}}Average \\ Levenshtein \\ similarity\end{tabular}} & \textbf{\begin{tabular}[c]{@{}l@{}}Average \\ sentiment \\ change\end{tabular}} \\ \hline
positive & 0.54 & 0.86 & 0.38 \\ \hline
neutral & 0.58 & 0.85 & 0.29 \\ \hline
negative & 0.53 & 0.86 & 0.42 \\ \hline
\end{tabular}
\caption{Performance comparision for different targets for style transfer. BERTScore threshold was set to 0.95 and maximum 10\% word perturbations were allowed. Search had a beam width of 3 and only transformation used was replacement.}
\label{table:polar}
\end{table}

As we can observe from table~\ref{table:polar}, using our proposed style transfer technique, we can obtain improved performance in transferring the style to positive or negative sentiment while maintaining comparable entailment and similarity values as in the case of neutrality change. 

\begin{figure}[h]
    \centering
    \subfloat[Target sentiment: Positive] {{\includegraphics[width=0.9\linewidth]{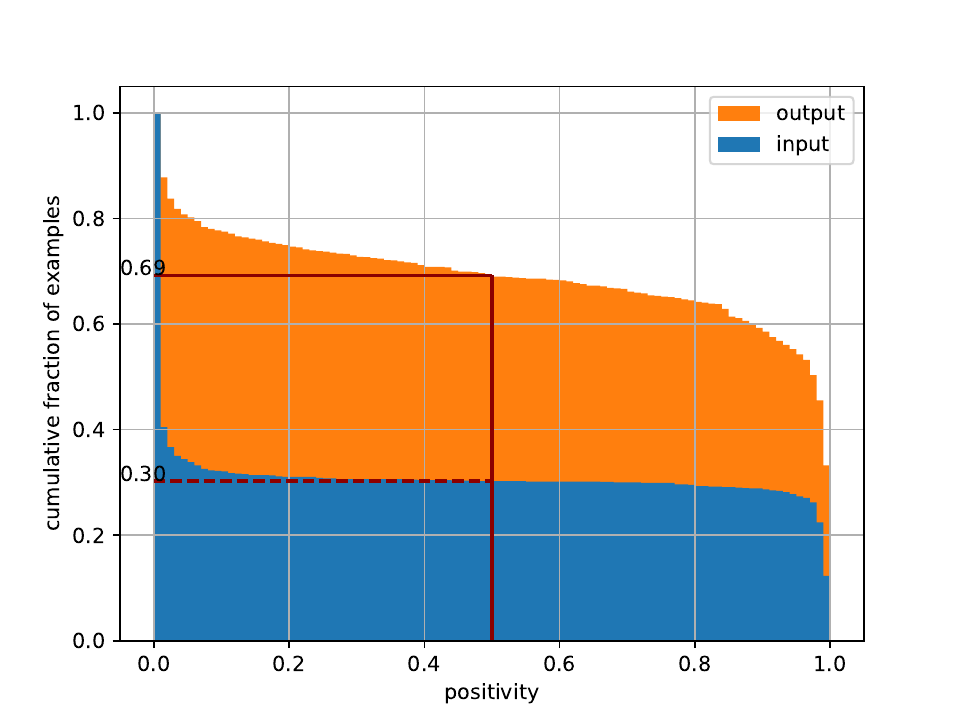}}}
    \hfill
    \subfloat[Target sentiment: Negative] {{\includegraphics[width=0.9\linewidth]{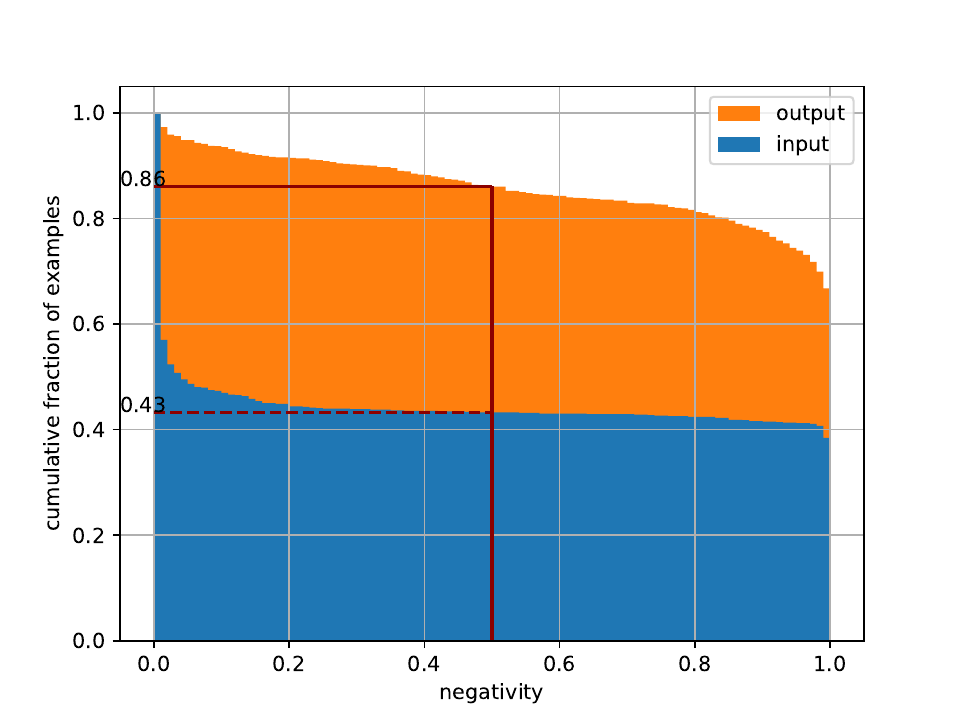}}}
    \caption{Cumulative fraction of examples vs polarity. Transformation: only replacement, min BERTScore:0.95, Beam width:3}
    \label{fig:cum_polar}
    \setlength{\belowcaptionskip}{10pt}
\end{figure}

In figure~\ref{fig:beam_cumulative}(b), we observed a change of 29\% in neutral classification of sentences. In case of positive classification, we observe a change of 39\% (figure~\ref{fig:cum_polar}(a)) and in case of negative classification, we observe a change of 43\% (figure~\ref{fig:cum_polar}(b)). This suggests doing style transfer to increase negativity in a sentence from our dataset for a particular aspect was easier as compared to increasing positivity or neutrality. Note that the input examples have a higher proportion of negative sentences as well.

\end{document}